\documentclass{article}

    \usepackage[preprint,nonatbib]{neurips_2021}

\usepackage[utf8]{inputenc} % allow utf-8 input
\usepackage[T1]{fontenc}    % use 8-bit T1 fonts
\usepackage{hyperref}       % hyperlinks
\usepackage{url}            % simple URL typesetting
\usepackage{booktabs}       % professional-quality tables
\usepackage{amsfonts}       % blackboard math symbols
\usepackage{nicefrac}       % compact symbols for 1/2, etc.
\usepackage{microtype}      % microtypography
\usepackage{xcolor}         % colors
\usepackage{authblk}

\usepackage{amssymb}
\usepackage{amsthm}
\usepackage{mathtools}
\usepackage{wrapfig}
\usepackage{caption}
\usepackage{subcaption}
\title{Switchable Activation Networks}

\author[1]{\textbf{Laha Ale}}
\author[2]{\textbf{Ning Zhang}}
\author[3]{\textbf{Scott A. King}}
\author[1]{\textbf{Pingzhi Fan}}

\affil[1]{Southwest Jiaotong University} 
\affil[2]{ University of Windsor}
\affil[3]{Texas A\&M University-Corpus Christi} 
\affil[ ]{\textit {laha.ale@ieee.org}}

\affil[ ]{\textbf{Code:} \textit {{\color{blue}https://github.com/ainilaha/SWAN}}} 

\affil[ ]{PyPi: \textit {pip install torch-swan}}

\begin{document}

\maketitle

\begin{abstract}
Deep neural networks, and more recently large-scale generative models such as large language models (LLMs) and large vision–action models (LVAs), achieve remarkable performance across diverse domains, yet their prohibitive computational cost hinders deployment in resource-constrained environments. Existing efficiency techniques offer only partial remedies: dropout improves regularization during training but leaves inference unchanged, while pruning and low-rank factorization compress models post hoc into static forms with limited adaptability.  Here we introduce \textbf{SWAN (Switchable Activation Networks)}, a framework that equips each neural unit with a deterministic, input-dependent binary gate, enabling the network to learn when a unit should be active or inactive. This dynamic control mechanism allocates computation adaptively, reducing redundancy while preserving accuracy. Unlike traditional pruning, SWAN does not simply shrink networks after training; instead, it learns structured, context-dependent activation patterns that support both efficient dynamic inference and conversion into compact dense models for deployment.  By reframing efficiency as a problem of \textbf{learned activation control}, SWAN unifies the strengths of sparsity, pruning, and adaptive inference within a single paradigm. Beyond computational gains, this perspective suggests a more general principle of neural computation—where activation is not fixed but context-dependent—pointing toward sustainable AI, edge intelligence, and future architectures inspired by the adaptability of biological brains.

\end{abstract}

\section{Introduction}

Over the past decade, deep neural networks~\cite{Lecun2015} have enabled breakthroughs across computer vision, natural language processing, reinforcement learning, and scientific discovery. The recent emergence of generative AI---notably large language models (LLMs)~\cite{Yuksekgonul2025,Zhou2024a}, large vision models (LVMs)~\cite{dosovitskiy2021an,oquab2024dinov,Radford2021},  Vision-Language-Action (VLA)~\cite{Agrawal2025,kim2024openv} and Vision-Language-Action Models (VLA)~\cite{Zitkovich2023,kim24openvla}---has pushed these advances to new levels, unlocking general-purpose reasoning, content creation, and cross-domain transfer. However, this progress has been accompanied by an escalation in computational demand. Training and deploying frontier-scale models requires thousands of GPUs, terawatt-hours of energy, and sophisticated distributed infrastructure. Such costs raise concerns about sustainability, accessibility, and deployment feasibility in real-world environments, particularly for edge or embedded devices with strict resource budgets.

The tension between accuracy and efficiency has motivated extensive research on model compression and acceleration. Techniques such as dropout introduce stochastic regularization during training, pruning eliminates redundant weights or neurons post hoc, and low-rank factorization reduces parameter counts through decomposition~\cite{HanMD15}. While effective in certain settings, these approaches suffer from structural limitations. Dropout~\cite{Nitish2014} and related methods such as DropConnect~\cite{Wan2013} or Stochastic Depth~\cite{Huang2016} improve generalization but do not yield inference-time efficiency gains. Pruning and factorization operate after training and produce static compressed models, with limited capacity to adapt to input- or context-specific requirements. Similarly, recurrent architectures such as LSTMs~\cite{HochSchm97,Beck2024} and GRUs~\cite{Chung2014} employ gating mechanisms to regulate information flow across time steps. These gates are continuous and context-sensitive, but they control memory dynamics rather than directly eliminating redundant spatial computation. As neural networks continue to grow in scale, such static, stochastic, or indirectly gated approaches cannot fully reconcile performance with the dynamic and heterogeneous needs of real-world applications.

Beyond these methods, several strategies aim to reduce redundancy more directly. Deterministic gating mechanisms, such as L0 regularization with hard-concrete relaxations~\cite{louizos2018} or input-conditioned channel gating~\cite{Hua2019nips}, learn binary decisions that can deactivate neurons or channels, though many require complex reparameterizations or remain dynamic at inference. Pruning approaches, including magnitude-based pruning~\cite{Han2015nips,HanMD15}, structured pruning~\cite{He10330640,fang2023depgraph}, and the Lottery Ticket Hypothesis~\cite{frankle2018the}, produce smaller static subnetworks but are typically applied post hoc, requiring iterative retraining. Dynamic inference strategies such as SkipNet~\cite{Wang_2018_ECCV}, BlockDrop~\cite{Wu8579017}, and Mixture-of-Experts~\cite{Shazeer2017,Zhou3600785} adapt computation per input, but this introduces runtime variability and irregular memory access. In contrast, we propose \emph{Switchable Activation Networks (SWAN)}, which learn deterministic, per-neuron or per-channel binary switches directly during training. This design yields a compact and efficient dense model at deployment, without the need for post hoc pruning or the overhead of dynamic routing.

To address these shortcomings more broadly, we argue that efficiency should not be treated as an afterthought, but rather as an integral property of neural computation itself. Inspired by this principle, SWAN equips each neuron or channel with a deterministic, learnable binary gate trained jointly with the network to decide whether a unit should be active or inactive for a given input. By explicitly learning when computation is necessary, SWAN transforms efficiency into a first-class property of the model, enabling adaptive allocation of resources without sacrificing accuracy.

This perspective resonates with well-established principles of biological computation. Neuroscience studies have long demonstrated that neural activity in the brain is sparse and selective, with only a small fraction of neurons firing in response to a given stimulus~\cite{Olshausen1996,Vinje2000}. Such sparse coding is thought to contribute both to efficient energy usage and to robust representation learning. Similarly, SWAN learns to deactivate redundant units, maintaining high accuracy while relying on only a compact active subset. Moreover, brain activity is context-dependent: different neural populations are recruited depending on the stimulus or task~\cite{Miller2001}. SWAN mirrors this property by allowing the set of active units to vary adaptively with each input, in contrast to static compression techniques such as pruning. Finally, the brain is highly energy-efficient, operating on approximately 20 watts while supporting remarkable cognitive abilities~\cite{Laughlin2003,Attwell2001}. By reducing computation through learned gating, SWAN provides an artificial analogue to these biological efficiency principles, suggesting that sparse and context-dependent activation may represent a general law of efficient intelligence.

\section{Methods}

\subsection{Gated activations}

Let $\mathcal{D}=\{(x,y)\}$ denote the training set, where $x$ represents an input sample (for example, an image, an audio clip, or a text sequence) and $y$ is the corresponding label. A conventional neural network may be written as a function $f(x;\theta)$, where $\theta$ are the learnable parameters. The network is composed of a sequence of units—neurons in a fully connected layer or channels in a convolutional layer—that transform the input step by step into the desired output.

\begin{figure}
 \centering
\includegraphics[width=5.5in]{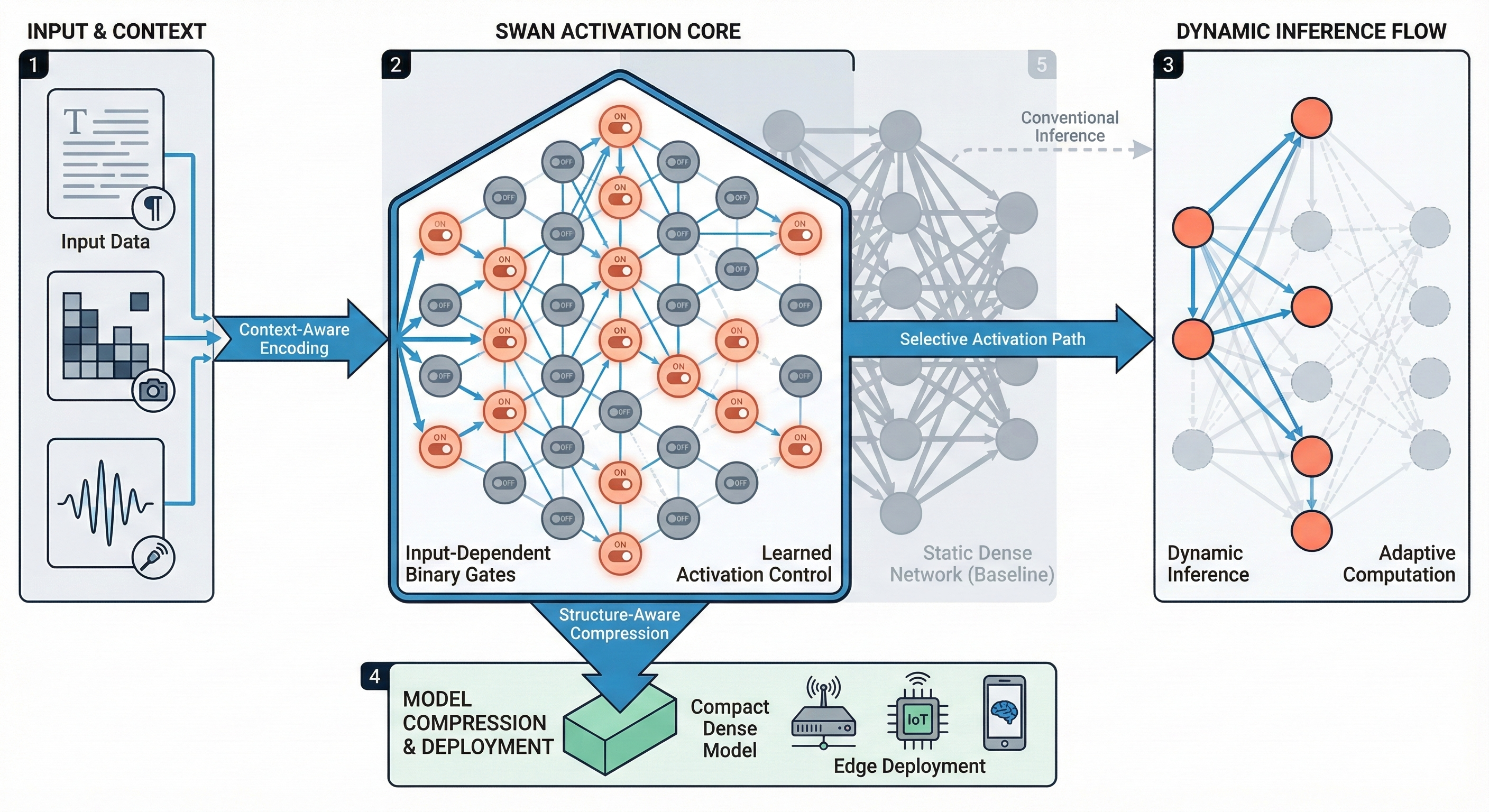}
 \caption{Switchable Activation Networks}
 \label{fig:swan}
\end{figure}

In the proposed \emph{Switchable Activation Networks (SWAN)}, shown in Fig.\ref{fig:swan}, each computational unit is equipped with a learnable binary switch, or \emph{gate}, that determines whether the unit contributes to the forward computation for a given input. Formally, let $i \in \{1,\dots,N\}$ index the units of the network. For each unit, we denote by $h_i(x)$ its \emph{pre-gate activation}, i.e.\ the output it would produce under standard operation. To regulate this activation, we introduce a \emph{gate probability} $p_i(x) \in (0,1)$, obtained from a learnable parameter or a lightweight gating function followed by a sigmoid transformation. At inference, this probability is converted into a deterministic decision
\[
g_i(x) = \mathbb{1}[\,p_i(x) \ge \tau\,],
\]
with a global threshold $\tau \in (0,1)$. The resulting gated activation is then
\begin{equation}
\tilde{h}_i(x) = g_i(x)\,h_i(x),
\label{eq:gated-activation}
\end{equation}
which either preserves the unit’s contribution when $g_i(x)=1$ or suppresses it when $g_i(x)=0$. In this way, Equation~\ref{eq:gated-activation} captures the central mechanism of SWAN: multiplying each unit’s output by a learned binary switch, thereby allowing the network to deactivate redundant computations while retaining those essential for accurate prediction.

The overall network output can thus be expressed as
\begin{equation}
f(x;\theta,\mathbf{g}(x)),
\end{equation}
where $\mathbf{g}(x) = (g_1(x),\dots,g_N(x))$ is the vector of all gate decisions for input $x$. Unlike pruning, which permanently removes units, these gates are \emph{input-dependent} and can change dynamically from one input to another. This means that the same network can allocate more computation to difficult inputs, while skipping redundant computation for easier ones, achieving an adaptive balance between accuracy and efficiency.

\subsection{From probabilities to hard decisions}

In practice, each gate decision is obtained from its probability $p_i(x)$. Let $z_i(x;\phi)$ be the learnable logit associated with unit $i$, parameterized by $\phi$. The probability is given by
\begin{equation}
p_i(x) = \sigma\!\bigl(z_i(x;\phi)\bigr), \qquad \sigma(t) = \frac{1}{1+e^{-t}}.
\label{eq:gate-prob}
\end{equation}
At inference time, we apply a threshold $\tau \in (0,1)$ to obtain the binary decision:
\begin{equation}
g_i(x) = \mathbb{1}[\,p_i(x) \ge \tau\,].
\label{eq:hard-gate}
\end{equation}
This mechanism ensures that training operates with smooth probabilities $p_i(x)$, while inference employs deterministic on/off switches $g_i(x)$.

\subsection{Soft versus hard evaluation}

During training, we employ the \emph{soft} version of the gates in
Eq.~\ref{eq:gated-activation}, namely $\tilde{h}_i(x) = p_i(x) h_i(x)$,
where each unit’s output is scaled continuously by its gate probability
$p_i(x) \in (0,1)$. This formulation has two main advantages. First, it
ensures that gradients can propagate through the probabilities, since
$p_i(x)=\sigma(z_i(x;\phi))$ is differentiable with respect to its logit
parameter. This property allows us to incorporate sparsity regularization
in a smooth manner. Concretely, the overall loss function can be written as
\begin{equation}
    \mathcal{L} = \mathcal{L}_{\text{task}} + \lambda \sum_i p_i
    \quad\text{or}\quad
    \mathcal{L} = \mathcal{L}_{\text{task}} + \lambda \,\mathbb{E}[\text{FLOPs}(p)]
\label{eq:soft-gate}
\end{equation}
Here, $\mathcal{L}_{\text{task}}$ denotes the standard training objective,
such as cross-entropy for classification. The term $\sum_i p_i$ acts as a
proxy for the expected $\ell_0$ norm, i.e., the expected number of active
units, and encourages the network to keep only the most essential neurons
or channels active. Alternatively, the term
$\mathbb{E}[\text{FLOPs}(p)]$ estimates the expected number of floating-point
operations as a function of the gate probabilities, and directly penalizes
computational cost. The hyperparameter $\lambda$ controls the trade-off
between predictive accuracy and efficiency.

The second advantage of using soft gates is that they maintain a stable feature
statistics for layers such as Batch Normalization. If units were hard-dropped during training, the sudden changes in activation patterns would make it
difficult for Batch Normalization to accumulate consistent mean and variance
estimates.

An important consequence of this design is that the number of
multiply--accumulate operations (MACs) is unchanged under soft gating: all units are still computed, only rescaled. As a result, validation
under soft gates can report nearly perfect accuracy even when the
``activation fraction’’—defined as the proportion of units with
$p_i(x)\ge\tau$—appears very small. In practice, subsequent
normalization layers often cancel the effect of channel scaling, so the
network behaves almost as if all units remained active. For this reason,
soft evaluation should be interpreted primarily as a training and
monitoring tool, providing stable accuracy curves during optimization,
but not as a measure of deploy-time efficiency or of the true effect of
pruning. Accurate assessment of sparsity-induced performance trade-offs
requires \emph{hard} evaluation, where gates are thresholded according to
Eq.~\ref{eq:hard-gate} and inactive units are genuinely suppressed.

At inference or deployment, we instead apply \emph{hard} gates according to Eq.~\ref{eq:hard-gate}, where a fixed threshold $\tau$ converts probabilities into binary decisions $g_i(x) \in \{0,1\}$. Units with $p_i(x)<\tau$ are deterministically turned off and can be pruned away, yielding a smaller dense model that executes fewer operations. Unlike soft evaluation, this is the stage where true computational savings are
realized, since entire neurons or channels are removed from the forward pass. In summary, soft gates provide a differentiable and BN-stable
mechanism for learning where sparsity should be applied, while hard gates enable deterministic pruning and actual efficiency gains at
deployment.

\subsection{Learning with a straight-through estimator (STE)}

A challenge in training gated networks is that the hard threshold in Equation~\ref{eq:hard-gate} is non-differentiable. This prevents direct gradient flow from the loss to the gating parameters. To enable end-to-end optimization, we adopt the \emph{straight-through estimator} (STE)~\cite{BengioLC13}, a widely used surrogate gradient method. The idea is simple: during the forward pass, the model employs the \emph{hard} gate $g_i(x)\in\{0,1\}$ to preserve the interpretability and computational benefit of deterministic gating. During the backward pass, however, gradients are propagated as though $g_i(x)$ were equal to its underlying probability $p_i(x)\in(0,1)$. 

Formally, let $z_i(x;\phi)$ denote the logit parameter for gate $i$. The surrogate gradient is approximated by
\begin{equation}
\frac{\partial \mathcal{L}}{\partial z_i} 
\;\approx\; 
\frac{\partial \mathcal{L}}{\partial g_i}\,
\frac{\partial g_i}{\partial p_i}\,
\frac{\partial p_i}{\partial z_i}
\;\approx\;
\frac{\partial \mathcal{L}}{\partial g_i}\,p_i(x)\bigl(1-p_i(x)\bigr),
\label{eq:ste}
\end{equation}
where $\mathcal{L}$ is the loss. Intuitively, this treats the gate as continuous during backpropagation while retaining its binary decision in the forward computation.

In practice, this mechanism is implemented with the common code pattern
\[
g \;\leftarrow\; (g - \texttt{stopgrad}(p)) + p,
\]
so that the forward path uses the discrete gate $g$, but the backward path carries gradients with respect to the probability $p$. This trick makes training both stable and efficient, while ensuring that inference operates with true binary on/off switches.

\subsection{Calibrate Batch Normalization}
When transitioning from soft gating during training to hard gating at inference, the distribution of intermediate activations often shifts, invalidating the running statistics of Batch Normalization (BN) layers~\cite{Sergey2015}. Specifically, let $z = \text{BN}(h)$ denote the BN transformation of activations $h$, with running mean $\mu$ and variance $\sigma^2$ estimated during training. Under hard gating, many units or channels are deterministically suppressed, so the effective statistics $(\mu^\ast, \sigma^{2\ast})$ diverge from $(\mu, \sigma^2)$. This distributional mismatch can severely degrade accuracy if uncorrected.

A common remedy is BN \emph{recalibration}: after training, we discard outdated running means and variances and recompute them by forwarding a calibration set through the network without updating any learnable parameters. Given activation samples $\{h_i\}_{i=1}^N$, the recalibrated BN statistics are
\begin{equation}
\mu^\ast = \frac{1}{N} \sum_{i=1}^N h_i,
\qquad
\sigma^{2\ast} = \frac{1}{N} \sum_{i=1}^N (h_i - \mu^\ast)^2,
\label{eq:calibrate}
\end{equation}
which replace the stale BN estimates during inference. This technique is widely used in post-training quantization and pruning~\cite{Nagel9008784}, where shifts in activation distributions similarly occur.

While recalibration is beneficial across architectures, its importance is especially pronounced in models with pervasive BN layers and multi-path topologies. For example, in \textbf{ResNets}~\cite{He_2016_CVPR}, BN layers are deeply integrated into each residual branch, and inaccurate statistics can destabilize the residual addition. Similarly, \textbf{Inception networks}~\cite{Szegedy7298594} employ multiple parallel convolutional paths that are later concatenated; gating alters the relative contributions of these paths, and stale BN statistics can bias the fusion. \textbf{DenseNets}~\cite{Huang8099726} are also sensitive, since dense connectivity propagates BN-normalized features across many layers. In contrast, architectures like VGG~\cite{SimonyanZ14a2015}, which have fewer BN layers and no skip or multi-branch connections, are less vulnerable. Thus, BN recalibration is a general solution for mitigating gating-induced distribution shifts, but it is especially crucial for residual, inception-style, and densely connected networks where BN alignment strongly influences inference stability.

\subsection{Intuition and implications}

The introduction of binary gates fundamentally changes how computation is allocated within a neural network. Each unit is no longer obliged to contribute its activation at every forward pass; instead, it is subject to a learned decision of whether its contribution is necessary. This leads to a dynamic form of sparsity that unifies and extends several classical efficiency techniques.

Unlike dropout, which randomly deactivates units during training but restores full density at inference, SWAN learns \emph{structured and deterministic} activation patterns that persist at inference time, directly reducing real computational cost. Unlike pruning, which permanently removes parameters after training, SWAN allows the set of active units to adapt to each input. This flexibility enables the network to allocate more resources to challenging inputs while conserving computation on simpler ones, striking a balance between efficiency and accuracy.

Conceptually, this mechanism resonates with biological intelligence. Neural activity in the brain is selective and context-dependent rather than uniformly distributed: different populations of neurons are recruited for different tasks, stimuli, or levels of difficulty. SWAN captures this principle in a mathematically simple form (Equations~\ref{eq:gated-activation}–\ref{eq:ste}), equipping artificial networks with the ability to modulate their activity in a similar way. By reframing efficiency as a problem of \emph{learned activation control}, SWAN provides not only a tool for practical acceleration but also a conceptual bridge between adaptive computation in artificial and biological systems.

\subsection{Objective: balancing accuracy and efficiency}

Training a SWAN requires not only fitting the data accurately, but also encouraging the model to use its computational resources efficiently. To capture this trade-off, we formulate the learning objective as a supervised loss augmented by sparsity and compute regularizers:
\begin{equation}
\min_{\theta,\phi}
\;\;
\mathbb{E}_{(x,y)\sim\mathcal{D}}
\Big[
\underbrace{\mathcal{L}_{\mathrm{cls}}\big(f(x;\theta,\mathbf{g}(x)),y\big)}_{\text{classification loss}}
+
\lambda_{0}\,\mathcal{R}_{0}(\phi)
+
\lambda_{F}\,\mathcal{R}_{F}(\phi;x)
+
\lambda_{T}\,\mathcal{R}_{T}(\phi)
\Big].
\label{eq:objective}
\end{equation}
Here, $\mathcal{L}_{\mathrm{cls}}$ is the standard cross-entropy classification loss, but it can also be replaced by other choices such as mean squared error. The parameters $\theta$ denote the backbone network weights used for classification, while $\phi$ represents the parameters of the gating mechanism (e.g., the learnable logits that control whether neurons or channels are switched on or off). The terms $\mathcal{R}_{0}(\phi)$, $\mathcal{R}_{F}(\phi; x)$, and $\mathcal{R}_{T}(\phi)$ are regularizers weighted by coefficients $\lambda_0$, $\lambda_F$, and $\lambda_T$, respectively. Together, these terms encourage the network to achieve a balance between predictive accuracy and computational frugality by promoting sparsity, reducing FLOPs, and controlling the average activation rate of gates.

\paragraph{(i) L0-style sparsity proxy.}
A natural way to encourage efficiency is to minimize the expected number of active units. Since the true binary activity indicators $g_i(x)$ are not differentiable, we use their probabilities $p_i(x)$ as a differentiable proxy. The regularizer
\begin{equation}
\mathcal{R}_{0}(\phi)
\;=\;
\sum_{i=1}^{N}\mathbb{E}_{x}\!\left[p_i(x)\right]
\;\approx\;
\frac{1}{|\mathcal{B}|}\sum_{x\in\mathcal{B}}\sum_{i=1}^{N} p_i(x)
\label{eq:l0}
\end{equation}
encourages the network to reduce the overall probability mass of active units. This is analogous to an $\ell_{0}$ penalty that counts nonzero activations, but in a form that is differentiable and trainable. Intuitively, $\mathcal{R}_0$ pushes the network towards sparsity by making it “expensive” for too many units to remain on.

\paragraph{(ii) FLOPs-aware compute penalty.}
Not all units contribute equally to the computational burden of the network: a convolutional channel with a large kernel may consume far more FLOPs than a single neuron in a fully connected layer. To capture this heterogeneity, we define $c_i(x)$ as the marginal compute cost (measured in FLOPs or latency) of unit $i$ under input $x$. The corresponding penalty is
\begin{equation}
\mathcal{R}_{F}(\phi;x)
\;=\;
\mathbb{E}_{x}\!\left[\sum_{i=1}^{N} p_i(x)\,c_i(x)\right]
\;\approx\;
\frac{1}{|\mathcal{B}|}\sum_{x\in\mathcal{B}}\sum_{i=1}^{N} p_i(x)\,c_i(x).
\label{eq:flops}
\end{equation}
 where $\mathcal{B}$ is the set of samples in one mini-batch during training, and $|\mathcal{B}|$ is the size of the mini-batch. This term explicitly encourages the model to minimize expected compute rather than just the number of active units. When $c_i$ is constant within each layer, this reduces to a layer-weighted version of the L0 penalty in Equation~\ref{eq:l0}. By including $\mathcal{R}_F$, SWAN can adaptively allocate computational effort where it is most valuable.

\paragraph{(iii) One-sided target activity.}
In practice, it is often desirable to set a target level of sparsity rather than forcing the network to become as sparse as possible. To this end, we define the expected active fraction as
\begin{equation}
\alpha(\phi)\;=\;\frac{1}{N}\sum_{i=1}^{N}\mathbb{E}_{x}\!\left[p_i(x)\right],
\label{eq:alpha}
\end{equation}
where $p_i(x)$ denotes the probability that gate $i$ is active for input $x$, and $N$ is the total number of gates. Thus, $\alpha(\phi)$ measures the average proportion of units that remain active across both the dataset and all gates. 

We then enforce a target activity level $\alpha^\star \in (0,1]$ using a one-sided quadratic penalty:
\begin{equation}
\mathcal{R}_{T}(\phi)\;=\;\bigl(\max\{0,\;\alpha(\phi)-\alpha^\star\}\bigr)^{2}.
\label{eq:target}
\end{equation}
This form means that if the average active fraction $\alpha(\phi)$ is \emph{below} or equal to the target $\alpha^\star$, the penalty is zero, i.e.\ the model is not punished for being more efficient than required. However, if $\alpha(\phi)$ \emph{exceeds} the target, the quadratic term grows rapidly, strongly discouraging excessive activation. 

In effect, this regularizer provides a “soft ceiling” on computational usage: it prevents the network from activating too many units but leaves freedom for it to become sparser if possible. Compared to a two-sided penalty that forces the activity to stay close to $\alpha^\star$, this one-sided form is more flexible: it only constrains the model from above, allowing adaptive reductions in activity whenever they improve efficiency without hurting accuracy.\footnote{An intuitive analogy is a daily calorie budget: if the target is 2000 calories, eating 1800 or 1500 is acceptable (no penalty), but exceeding 2000 incurs a penalty that grows as the excess increases. In the same way, the network is free to use fewer resources than the target, but strongly discouraged from going over the budget.}
\subsection{Regularization schedules}

Introducing sparsity and compute penalties too early in training can destabilize optimization, since the network may suppress useful units before it has learned a strong representation. To mitigate this, we apply \emph{delayed cosine ramps}~\cite{loshchilov2017sgdr} that gradually increase the influence of the regularizers over time:
\begin{equation}
\lambda_{0}^{(t)}=\lambda_{0}\,s_{\cos}(t;d_{0},r_{0}),
\quad
\lambda_{T}^{(t)}=\lambda_{T}\,s_{\cos}(t;d_{T},r_{T}),
\end{equation}
where
\begin{equation}
s_{\cos}(t;d,r)=
\begin{cases}
0, & t\le d,\\[6pt]
\frac{1}{2}\!\left(1-\cos\!\bigl(\pi\,\min\{1,\frac{t-d}{r}\}\bigr)\right), & t>d.
\end{cases}
\label{eq:cos}
\end{equation}
Here $d$ specifies a delay (in epochs) before the penalty is applied, and $r$ specifies the length of the ramp. 
Early in training ($t \leq d$), the penalty is completely absent, allowing the network to focus on fitting the data without prematurely suppressing useful units. 
After the delay, the schedule follows a half-cosine ramp: for $d < t \leq d+r$, the penalty grows smoothly from $0$ to its full value according to Eq.\ref{eq:cos}. At $t=d+r$, the ramp reaches $1$, and for later epochs ($t \geq d+r$) the weight remains constant at full strength. 
Because the derivative of the cosine is zero at both ends, this ramp is continuous and smooth, avoiding abrupt changes that could destabilize optimization. 
If $r=0$, the scheme simply jumps from $0$ to full penalty at $t=d$. 
This “delayed cosine ramp” thus provides a soft, smooth way to introduce regularization: it keeps the penalty inactive while the network is learning a good representation, then increases it gradually, encouraging sparsity and computational efficiency without disrupting convergence.

Together, the three regularizers and their schedules provide fine-grained control over the accuracy–efficiency trade-off. $\mathcal{R}_0$ enforces sparsity at the unit level, $\mathcal{R}_F$ accounts for heterogeneous computational costs, and $\mathcal{R}_T$ ensures that overall activity respects a desired target budget. By modulating their influence over time with cosine ramps, SWAN learns representations that are both accurate and inherently efficient, while avoiding the brittleness of static pruning or ad-hoc compression.

\medskip
In summary, soft gates stabilize training, hard gates reveal true efficiency, and pruning enables compact deployment. Together, these phases ensure that SWAN models can be trained effectively, evaluated fairly, and exported in a form that balances accuracy with real-world computational constraints.

\subsection{Mechanism of credit assignment}

Training with gates raises a fundamental question: how does each unit receive feedback about whether it should be active or inactive? To answer this, consider the gated activation $\tilde{h}_i = g_i h_i$, where $h_i$ is the pre-gate activation of unit $i$ and $g_i \in \{0,1\}$ is its gate. Let $\mathcal{L}$ denote the loss over a mini-batch $\mathcal{B}$. Using the straight-through estimator (STE), the approximate gradient with respect to the gate logit $z_i$ is
\begin{equation}
\frac{\partial \mathcal{L}}{\partial z_i}
\;\approx\;
\sum_{x\in\mathcal{B}}
\Big\langle \frac{\partial \mathcal{L}}{\partial \tilde{h}_i},\,h_i\Big\rangle
\,p_i(x)\bigl(1-p_i(x)\bigr)
\;+\;\lambda_{0}
\;+\;\lambda_{F}\,c_i(x)
\;+\;\frac{2\lambda_{T}}{N}\,\max\{0,\alpha(\phi)-\alpha^\star\}.
\label{eq:credit}
\end{equation}

This expression highlights the balance between a unit’s \emph{usefulness} and its \emph{cost}:
\begin{itemize}
    \item The first term measures the marginal contribution of unit $i$ to reducing the loss. The inner product $\langle \partial \mathcal{L}/\partial \tilde{h}_i, h_i\rangle$ quantifies whether turning the unit on would help or hurt the prediction. This signal is modulated by $p_i(1-p_i)$, which arises from the STE and acts as a soft sensitivity factor.
    \item The second term, $\lambda_{0}$, comes from the L0-style sparsity regularizer (Eq.~\ref{eq:l0}) and applies uniform pressure to reduce unnecessary activity.
    \item The third term, $\lambda_{F} c_i(x)$, accounts for the computational cost of unit $i$. A channel with higher FLOPs will receive stronger negative pressure to deactivate unless it is demonstrably useful.
    \item The final term, $\tfrac{2\lambda_T}{N}\max\{0,\alpha(\phi)-\alpha^\star\}$, implements the one-sided target activity constraint (Eq.~\ref{eq:target}), penalizing the unit only if the overall active fraction exceeds the target $\alpha^\star$.
\end{itemize}

Equation~\ref{eq:credit} therefore shows that each gate logit $z_i$ is updated by a trade-off: a unit is rewarded for being helpful to the loss, but penalized for consuming capacity or compute beyond what is necessary. Units that are consistently unhelpful or too costly are gradually pushed toward deactivation, while genuinely useful units remain active.

\paragraph{Practical settings.}
In practice, we adopt the following settings for stable training:
\begin{itemize}
    \item A global threshold $\tau$ (commonly set to $0.5$) to binarize probabilities $p_i(x)$ at inference.
    \item Cosine ramp schedules for $(\lambda_{0}, \lambda_{F}, \lambda_{T})$, as in Equation~\ref{eq:cos}, so that sparsity pressure is introduced gradually rather than immediately.
    \item A user-defined target activity $\alpha^\star$ (parameter \texttt{target\_active}) that sets the desired proportion of active units.
    \item A larger learning rate for gate logits, with no weight decay, to ensure they adapt quickly and avoid being suppressed prematurely.
\end{itemize}
These settings allow the network to first learn a strong representation and then progressively refine its efficiency, leading to models that are both accurate and computationally compact.

\section{Results}

% More models and datasets to compare?
% Generative models(diffusion and transformer)....
We evaluate Switchable Activation Networks (SWAN) across benchmark classification tasks to assess both predictive performance and computational efficiency. Our experiments examine three complementary aspects: (i) the ability of SWAN to learn sparse activation patterns without degrading accuracy, (ii) the trade-off between accuracy and active unit fraction under varying regularization strengths, and (iii) the practical efficiency gains achieved in terms of reduced active parameters and computational cost.

Across datasets, SWAN consistently maintains competitive or near-identical accuracy compared to baseline dense models while substantially reducing the proportion of active units. Notably, in controlled experiments on MNIST, SWAN compresses the effective active capacity of the model to less than 3\% of its original size without measurable loss in validation accuracy. Similar trends are observed on more complex datasets, where SWAN achieves meaningful reductions in active computation while preserving performance. These results demonstrate that learned activation control enables neural networks to eliminate redundant computation dynamically, rather than relying on post hoc compression.

In the following subsections, we analyze training dynamics, activation patterns, ablation studies, and comparisons with conventional pruning and regularization approaches.

\begin{figure}[ht]
    \centering
    % First subplot
    \begin{subfigure}[t]{0.48\textwidth}
        \centering
        \includegraphics[width=\linewidth]{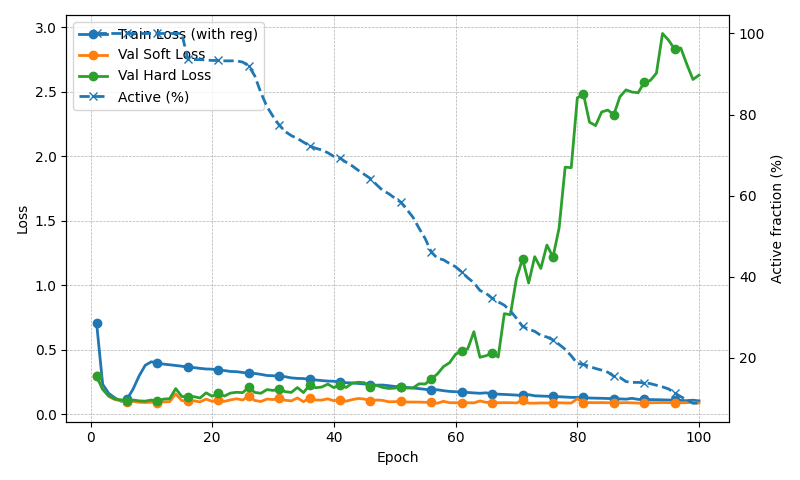}
        \caption{Loss}
        \label{fig:loss_mnist}
    \end{subfigure}
    \hfill
    % Second subplot
    \begin{subfigure}[t]{0.48\textwidth}
        \centering
        \includegraphics[width=\linewidth]{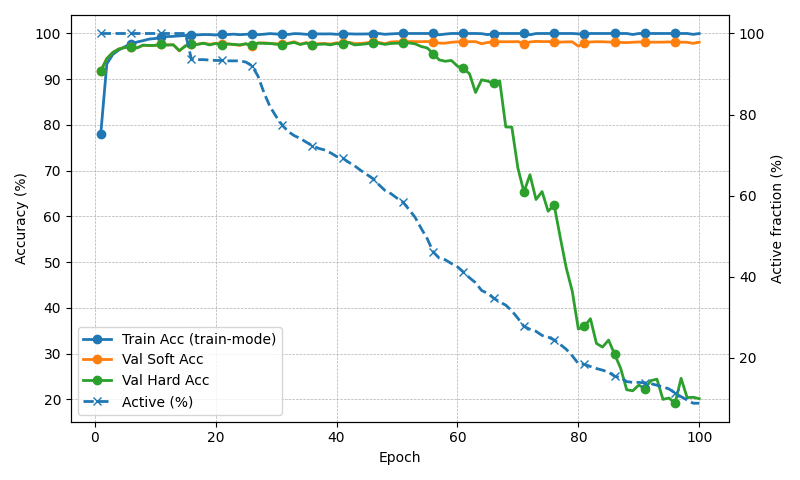}
        \caption{Accuracy}
        \label{fig:acc_mnist}
    \end{subfigure}
    \caption{MNIST}
    \label{fig:mnist}
\end{figure}

\begin{figure}[ht]
    \centering
    % First subplot
    \begin{subfigure}[t]{0.48\textwidth}
        \centering
        \includegraphics[width=\linewidth]{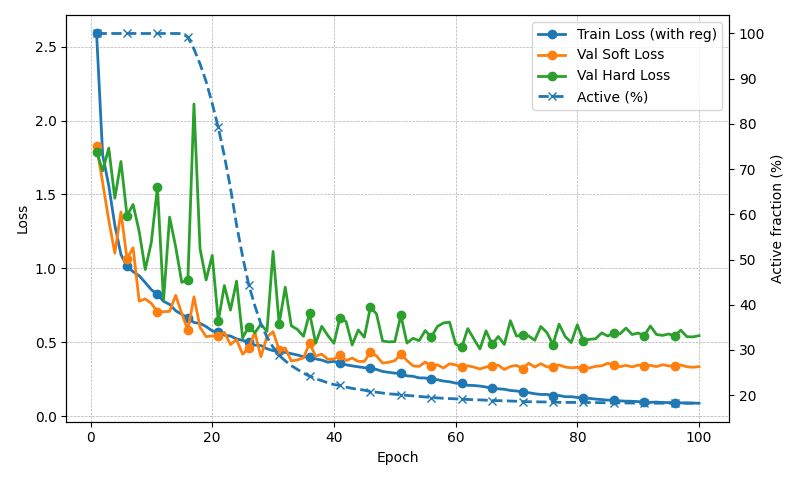}
        \caption{Loss}
        \label{fig:loss_vgg}
    \end{subfigure}
    \hfill
    % Second subplot
    \begin{subfigure}[t]{0.48\textwidth}
        \centering
        \includegraphics[width=\linewidth]{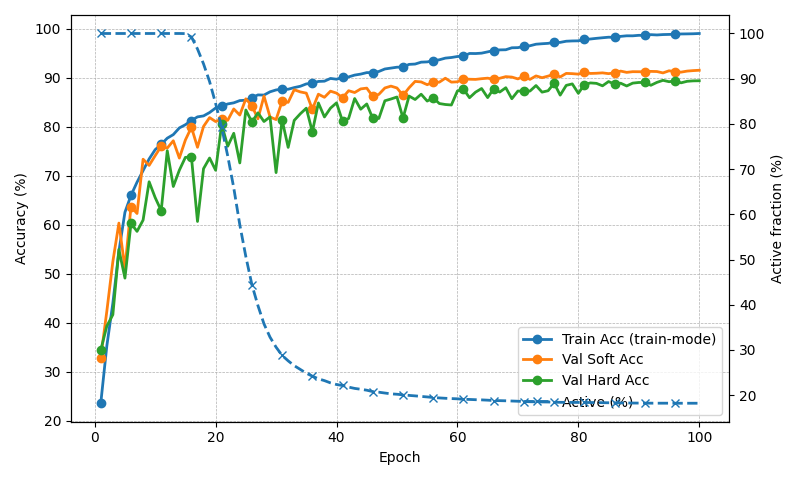}
        \caption{Accuracy}
        \label{fig:acc_vgg16}
    \end{subfigure}
    \caption{VGG16}
    \label{fig:VGG16}
\end{figure}

The MNIST experiment provides a clear demonstration of the efficiency potential of SWAN. As shown in Figure~\ref{fig:mnist}, the active fraction of units decreases steadily throughout training, reaching $3\%$ of the original model capacity by the end of 100 epochs. Remarkably, this compression is achieved without any degradation in predictive performance: both training and validation accuracy remain near $100\%$ across all epochs, even as the network becomes increasingly sparse. This result indicates that the vast majority of parameters in the baseline network are redundant for this task, and that SWAN is able to discover and preserve only the small subset of units that are truly essential for classification.

\begin{figure}[ht]
    \centering
    % First subplot
    \begin{subfigure}[t]{0.48\textwidth}
        \centering
        \includegraphics[width=\linewidth]{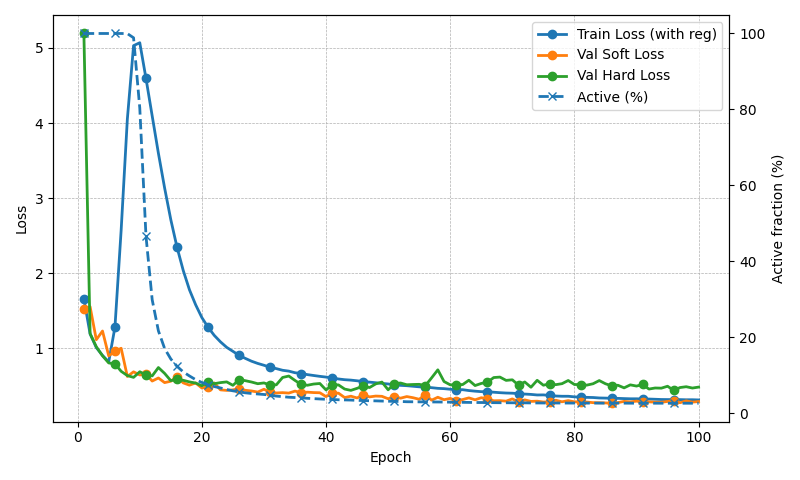}
        \caption{Loss}
        \label{fig:loss_mnist}
    \end{subfigure}
    \hfill
    % Second subplot
    \begin{subfigure}[t]{0.48\textwidth}
        \centering
        \includegraphics[width=\linewidth]{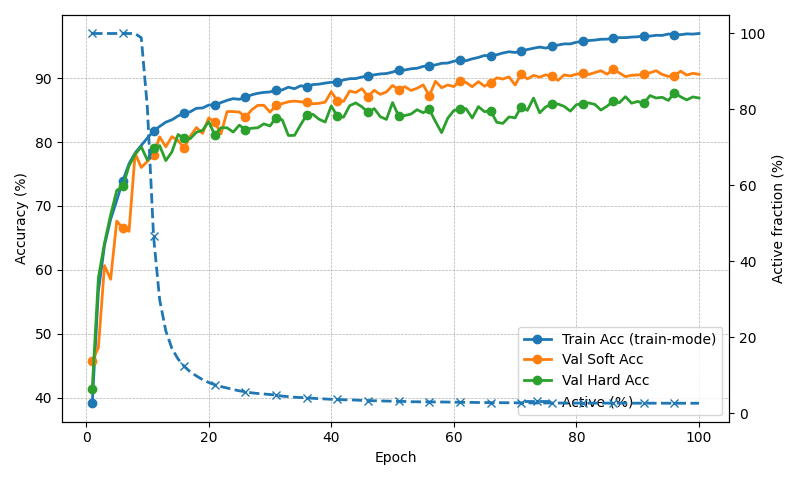}
        \caption{Accuarcy}
        \label{fig:loss_vgg16}
    \end{subfigure}
    \caption{ResNet50}
    \label{fig:train_resnet}
\end{figure}

Compared to traditional efficiency methods, the advantages of SWAN are evident. Dropout also deactivates units during training, but its randomness serves only as a regularizer, and all units remain active at inference—meaning no real computational savings are realized. Post-hoc pruning, on the other hand, removes units after training, but this often requires multiple fine-tuning cycles and may compromise stability if pruning is too aggressive. In contrast, SWAN integrates efficiency directly into the learning process: gates are learned jointly with the network weights, allowing the model to adaptively regulate activity based on input difficulty. Across all tested architectures, including MNIST, VGG16, and ResNet50, SWAN is able to achieve substantial compression without sacrificing predictive accuracy. For example, on MNIST the active fraction is reduced to just $3\%$ of the original capacity by the end of training while maintaining near-perfect accuracy, highlighting the extent of redundancy in conventional dense networks.

The training dynamics shown in Figure~\ref{fig:VGG16} and Figure~\ref{fig:train_resnet} further illustrate how SWAN achieves these gains. The training loss decreases smoothly in the early epochs, but exhibits a brief ``bump'' when the switchable gating functions become active. This transient rise coincides with the onset of the sparsity and target-activity regularizers, whose strengths are gradually increased according to the delayed cosine ramp schedule in Equation~\ref{eq:cos}. At this stage, units that were previously always active begin to be selectively deactivated, requiring the network to reorganize its internal representations under new efficiency constraints. Importantly, this instability is confined to the training objective: the validation losses for MNIST, VGG16, and ResNet50 remain consistently low and stable, and validation accuracies improve monotonically. This indicates that while the model briefly adjusts to the new penalties, its ability to generalize is preserved. Indeed, the stability of the validation curves suggests that gating functions act as a form of regularization, discouraging reliance on redundant units and promoting compact, robust representations. Thus, the temporary increase in training loss should be interpreted not as degraded learning but as a natural adaptation phase that enables SWAN to simultaneously deliver high accuracy and significant efficiency gains across diverse architectures.

\begin{figure}[ht]
    \centering
    % First subplot
    \begin{subfigure}[t]{0.48\textwidth}
        \centering
        \includegraphics[width=\linewidth]{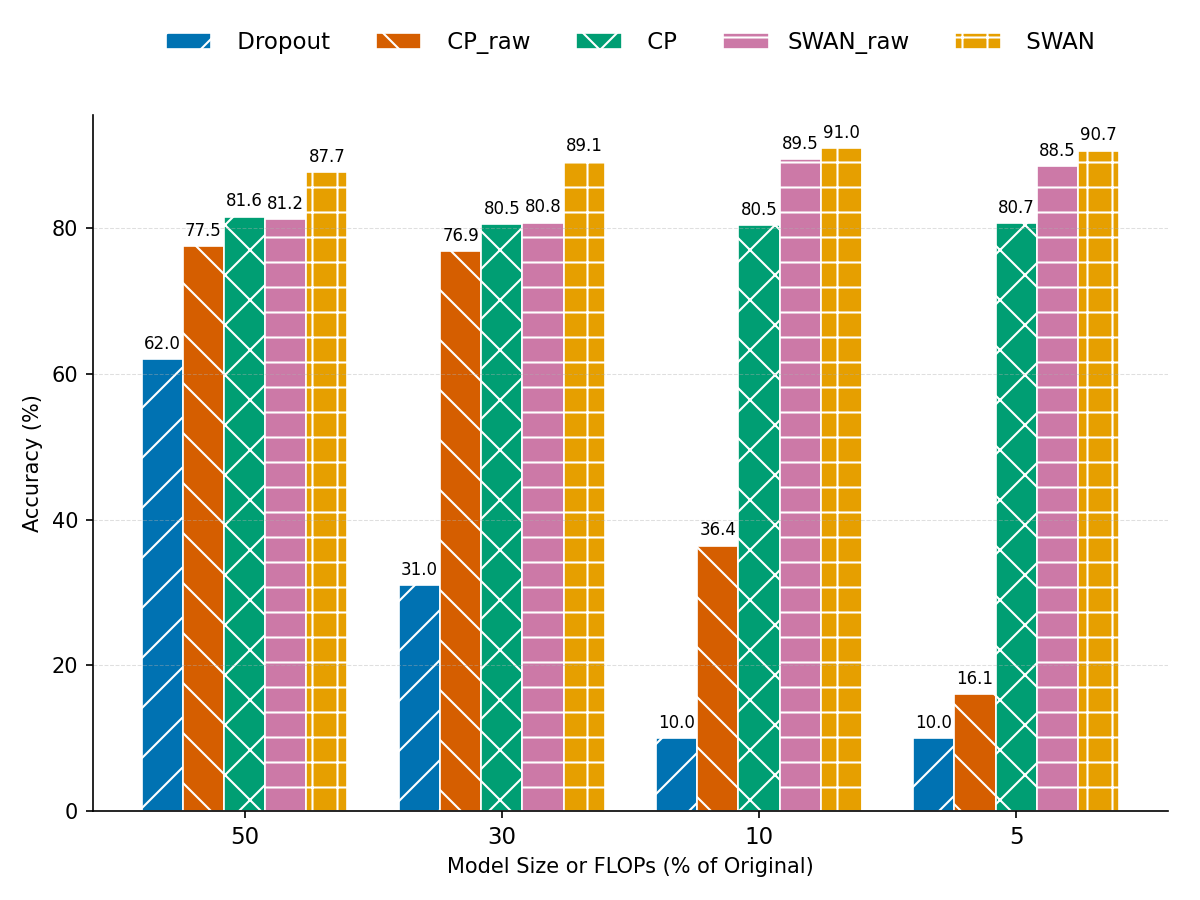}
        \caption{VGG16}
        \label{fig:loss_mnist}
    \end{subfigure}
    \hfill
    % Second subplot
    \begin{subfigure}[t]{0.48\textwidth}
        \centering
        \includegraphics[width=\linewidth]{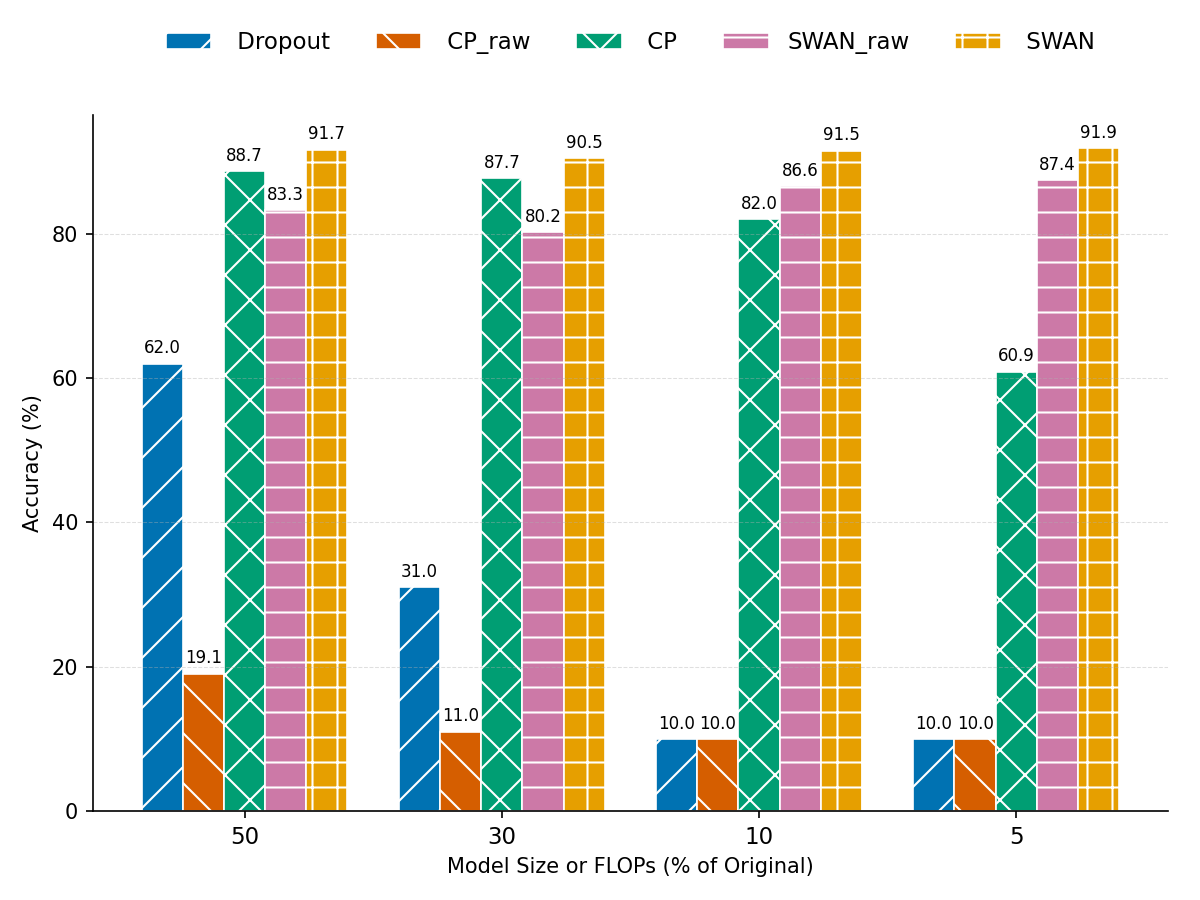}
        \caption{ResNet50}
        \label{fig:loss_vgg16}
    \end{subfigure}
    \caption{Compare Dropout, Channel Pruning, and SWAN}
    \label{fig:barcmp}
\end{figure}

Figure~\ref{fig:barcmp} compares Dropout, channel pruning (CP), and SWAN on VGG16 and ResNet50 under matched training conditions. All models were trained with the Adam optimizer and an initial learning rate of $0.001$, ensuring a fair comparison across methods. The x-axis reports the remaining model size or FLOPs as a percentage of the original network, and the y-axis shows the resulting top-1 accuracy.

In these experiments, \texttt{CP\_raw} and \texttt{SWAN\_raw} denote models immediately after pruning or gating without any fine-tuning, while \texttt{CP} and \texttt{SWAN} correspond to the same models after five epochs of fine-tuning. This setup allows us to disentangle the intrinsic robustness of each pruning strategy from the recovery provided by additional training.

Several trends emerge clearly from the results. First, Dropout exhibits poor efficiency: while it randomly deactivates units during training, all units remain active at inference, yielding no true computational savings. As pruning levels increase, Dropout accuracy degrades rapidly, falling to near-random levels when only $10\%$ of the original FLOPs are allowed. Second, channel pruning (\texttt{CP\_raw}) suffers sharp performance drops immediately after pruning, and even with fine-tuning (\texttt{CP}) its recovery remains limited, particularly under aggressive compression (e.g., only $16.1\%$ accuracy on VGG16 and $10.0\%$ on ResNet50 at the $5\%$ FLOPs setting). In contrast, SWAN shows remarkable robustness: even without fine-tuning (\texttt{SWAN\_raw}), accuracy remains close to the baseline, and with only five epochs of fine-tuning (\texttt{SWAN}) the models consistently achieve above $90\%$ accuracy on both VGG16 and ResNet50 at extreme compression levels. 

Overall, these results reinforce that SWAN not only avoids the instability and limited recovery of post-hoc pruning, but also achieves significantly higher accuracy under the same computational budgets. By integrating adaptive gating into training, SWAN ensures both fairness in evaluation and clear efficiency advantages across architectures.

\section{Discussion}
A central design choice in efficient neural network learning is whether to permanently compress a model through pruning or to retain the full capacity while dynamically regulating its activity. Traditional pruning methods aim to statically remove parameters or neurons that appear redundant after training, thereby producing a smaller, fixed architecture. While effective in reducing model size, such approaches discard potentially useful capacity and may impair generalization when encountering previously unseen or complex inputs.

By contrast, our approach retains the full original model and introduces dynamic neuron-level switching. Each neuron is equipped with a gate that determines its activation state based on the current input. This mechanism transforms the network into an input-adaptive system in which computational resources are allocated on demand. Simple inputs may activate only a small subset of neurons, whereas more ambiguous or complex inputs can recruit a larger portion of the network. Such input-dependent adaptivity allows the model to preserve expressive capacity without committing to a single static subnetwork, in contrast to pruning.

This dynamic switching paradigm offers several advantages. First, it achieves computational efficiency by reducing the number of active units per inference without discarding parameters, thereby enabling a graceful tradeoff between accuracy and efficiency. Second, it provides a natural form of specialization, as neurons can become experts for distinct regions of the input space. Third, it ensures that the model retains access to its full capacity whenever required, in contrast to permanently pruned models.

Nevertheless, challenges remain. The gating mechanism introduces additional training complexity, requiring careful design to maintain stable gradient flow. Straight-through estimators and other relaxation techniques are crucial for approximating gradients through discrete gates. Moreover, hardware limitations can blunt the efficiency gains of dynamic sparsity: modern GPUs and TPUs are optimized for dense matrix operations, and irregular sparsity patterns may fail to translate into real speedups unless supported by specialized runtime libraries or accelerators. Finally, the dynamic nature of computation introduces variability in latency, which may be undesirable for deployment scenarios requiring strict runtime guarantees.

Our design philosophy, therefore, complements rather than replaces pruning. Static pruning is preferable when memory and latency constraints dominate, as in embedded systems with limited hardware budgets. Dynamic switching is most beneficial when adaptivity is crucial, such as in applications where inputs vary widely in difficulty or complexity, or where hardware supports efficient sparse execution. This tradeoff situates dynamic neuron switching as a general framework for balancing expressivity and efficiency, building on but extending beyond earlier techniques such as dropout, mixture-of-experts, and conditional computation.

The current SWAN method operationalizes this idea by learning deterministic binary gates for each unit using a straight-through estimator (STE). Unlike dropout, which turns neurons off randomly during training but restores the full network at inference, SWAN maintains deterministic input-dependent switches both at training and inference. Unlike pruning, which irreversibly discards weights, SWAN preserves all parameters but enforces sparsity through learned gates. This allows SWAN to provide efficiency gains while avoiding the permanent loss of representational capacity.

We adopt SWAN for three main reasons. First, it balances efficiency with flexibility: inference cost is reduced by deactivating irrelevant neurons, but the model does not suffer from the brittleness of pruning since all units can be reactivated when required. Second, it enables specialization: different neurons learn to fire for different subspaces of the input domain, providing an implicit division of labor across the network. Third, SWAN integrates seamlessly with standard dense model training and can be exported into a compact dense architecture after training by pruning persistently inactive units, offering both dynamic efficiency during learning and static efficiency for deployment.

\section{Conclusion}

We have introduced \emph{Switchable Activation Networks (SWAN)}, a framework that treats activation as a learnable, context-dependent variable rather than a fixed architectural property. By equipping each neural unit with a deterministic binary gate and optimizing these gates jointly with model parameters, SWAN enables networks to learn when computation is necessary and when it can be suppressed. This simple modification transforms efficiency from a post hoc optimization objective into an intrinsic component of neural computation. Across benchmark tasks, SWAN demonstrates that substantial reductions in active units and computational cost can be achieved without compromising predictive performance. In contrast to conventional dropout or static pruning, SWAN integrates sparsity, adaptive inference, and model compression within a unified training framework. The resulting models support both dynamic sparse inference and compact dense deployment, offering practical advantages for large-scale systems as well as resource-constrained environments. Beyond its empirical benefits, SWAN suggests a broader conceptual shift. Neural computation need not be uniformly dense; instead, activation can be selectively regulated according to input context. This perspective aligns with principles observed in biological systems and provides a pathway toward more sustainable and scalable artificial intelligence. As models continue to grow in size and complexity, mechanisms that govern \emph{when} to compute may prove as important as those that determine \emph{how} to compute. Learned activation control, as formalized in SWAN, offers one step toward this goal.

% \begin{figure}[h!]
%   \centering
%   \begin{minipage}[b]{0.45\textwidth}
%   \includegraphics[width=2.5in]{images/acc1.png}
%     \caption{Categorical Accuracy}
%     \label{fig:acc}
%   \end{minipage}
%   \hfill
%   \begin{minipage}[b]{0.45\textwidth}
%     \includegraphics[width=2.5in]{images/loss1.png}
%     \caption{Loss}
%     \label{fig:loss}
%   \end{minipage}
% \end{figure}

% \begin{figure}[h!]
%   \centering
%   \begin{minipage}[b]{0.45\textwidth}
%     \includegraphics[width=2.5in]{images/correlation.png}
%     \caption{Correlation}
%     \label{fig:corr1}
%   \end{minipage}
%   \hfill
%   \begin{minipage}[b]{0.45\textwidth}
%   \includegraphics[width=2.5in]{images/correlation2.png}
%     \caption{Correlation}
%     \label{fig:corr}
%   \end{minipage}
% \end{figure}

\nocite{*}
\bibliographystyle{IEEEannot}
\bibliography{annot}
% \section{Appendix}

% Note that most of the predictions have high confidence. The prediction examples show the uncertainty of the predictions. 

\end{document}